\documentclass{article}
\usepackage{arxiv}
\pdfoutput=1

\usepackage[utf8]{inputenc} 
\usepackage[T1]{fontenc}    
\usepackage{hyperref}       
\usepackage{url}            
\usepackage{booktabs}       
\usepackage{amsfonts}       
\usepackage{nicefrac}       
\usepackage{microtype}      
\usepackage{lipsum}
\usepackage{graphicx}

\usepackage{amsmath}
\usepackage{subcaption}
\usepackage{rotating}
\usepackage{float}
\usepackage{adjustbox} 
\usepackage{multirow}
\usepackage{xcolor}

\title{Predict Confidently, Predict Right: Abstention in Dynamic Graph Learning}

\author{
 Jayadratha Gayen \\
  Machine Learning Lab\\
  IIIT Hyderabad\\
  \texttt{jayadratha.gayen@research.iiit.ac.in} \\
  \And
 Himanshu Pal \\
  Machine Learning Lab\\
  IIIT Hyderabad\\
  \texttt{himanshu.pal@research.iiit.ac.in} \\
  \And
 Naresh Manwani \\
  Machine Learning Lab\\
  IIIT Hyderabad\\
  \texttt{naresh.manwani@iiit.ac.in} \\
  \And
 Charu Sharma \\
  Machine Learning Lab\\
  IIIT Hyderabad\\
  \texttt{charu.sharma@iiit.ac.in} \\
}

\begin{document}

\maketitle
\begin{abstract}
Many real-world systems can be modeled as dynamic graphs, where nodes and edges evolve over time, requiring specialized models to capture their evolving dynamics in risk-sensitive applications effectively. Temporal graph neural networks (GNNs) are one such category of specialized models. For the first time, our approach integrates a reject option strategy within the framework of GNNs for continuous-time dynamic graphs. This allows the model to strategically abstain from making predictions when the uncertainty is high and confidence is low, thus minimizing the risk of critical misclassification and enhancing the results and reliability. We propose a coverage-based abstention prediction model to implement the reject option that maximizes prediction within a specified coverage. It improves the prediction score for link prediction and node classification tasks. Temporal GNNs deal with extremely skewed datasets for the next state prediction or node classification task. In the case of class imbalance, our method can be further tuned to provide a higher weightage to the minority class. Exhaustive experiments are presented on four datasets for dynamic link prediction and two datasets for dynamic node classification tasks. This demonstrates the effectiveness of our approach in improving the reliability and area under the curve (AUC)/ average precision (AP) scores for predictions in dynamic graph scenarios. The results highlight our model's ability to efficiently handle the trade-offs between prediction confidence and coverage, making it a dependable solution for applications requiring high precision in dynamic and uncertain environments.
\end{abstract}


\section{Introduction}
In the modern era, numerous systems are modeled as dynamic graphs where nodes and edges evolve over time. These systems include social and interaction networks~\cite{DBLP:conf/kdd/KumarZL19}, traffic networks \cite{DBLP:conf/nips/0001YL0020} trade networks \cite{poursafaei2022towards}, biological networks\cite{behrouz2023learning} and transaction network\cite{nath2023tboost} among others. Particularly in risk-sensitive applications such as fraud detection\cite{reddy2021tegraf}, fake news, polarization identification\cite{chomel2022polarization}, financial transaction\cite{nath2023tboost}, blockchain security \cite{behrouz2022anomaly}, epidemic modeling\cite{varugunda2023exploring, nguyen2023predicting}, anomaly detection\cite{reha2023anomaly} and disease prediction\cite{behrouz2022anomaly}, the stakes for accurate and reliable predictions are exceptionally high. 

Traditional graph neural networks (GNNs) and their extensions to dynamic graphs have shown promise in capturing the complex interactions and evolving structures within these networks. However, they often fall short where the cost of misclassification is significant and abstaining from making a prediction could mitigate risk.
\begin{figure}[t]
\centering
    \includegraphics[width=0.65\linewidth, trim=2cm 1.1cm 2cm 1.1cm, clip]{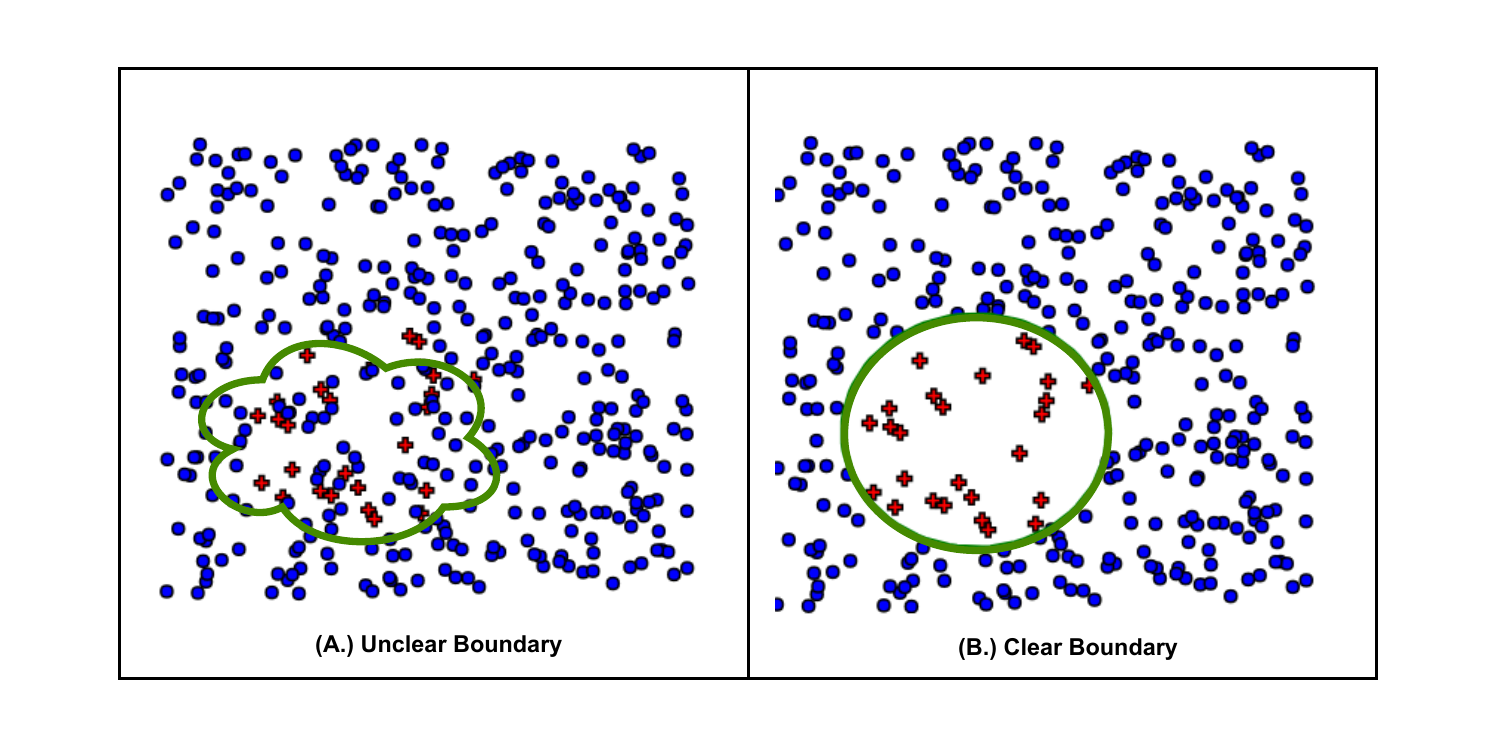}
    \caption{Decision Boundary: (A) is not clear and smooth (B) becomes clearer after abstaining confusing/noisy samples. The examples to be rejected are decided by the objective function, which consists of the losses over unrejected samples and a fixed cost for rejected samples. Minimum value of the objective will be achieved only with proper choice of rejected examples.}
    \label{fig:teaser}
\end{figure}
The additional challenge of highly skewed temporal\footnote{``temporal'' and ``dynamic'' will be used interchangeably.} datasets for node classification (such as Wikipedia, Reddit datasets\cite{DBLP:conf/kdd/KumarZL19}) makes it harder for the models to predict correctly. Recent works on temporal GNNs\cite{DBLP:journals/corr/abs-2006-10637, cong2023do, 
DBLP:journals/corr/abs-2101-05974,souza2022provably,DBLP:conf/iclr/XuRKKA20, DBLP:conf/kdd/KumarZL19} solve for node classification and link prediction tasks. 

In many fraud or anomaly detection problems, anomalous events are infrequent and most events are genuine only. Figure~\ref{fig:teaser}(A) illustrates such a classification problem with high class imbalance. Moreover, minority class points are completely overlapping with the class region of the majority class. Classifiers which can achieve very high accuracy on such data become very complex. The objective in such problems is to build a classifier which can achieve very low false negative rate on anomalous events with very small increase in false positive rate. A simpler approach would be removing majority class points from the overlapping region and learning the classification boundary. Figure~\ref{fig:teaser}(B) shows that the classifier required after abstaining some majority points from the overlapping region is much simpler. It also achieves a very low false negative rate on minority class with some increase in false positive rate. 

Classifiers with abstention are well explored in machine learning community\cite{manwani2015double,geifman2019selectivenet,pmlr-v161-kalra21a,pmlr-v139-charoenphakdee21a,cao2022generalizing}. Such models allow a classifier to withhold a prediction when the uncertainty associated with the decision is high. This approach has not been widely investigated in the context of dynamic graphs, where the temporal evolution of the data adds a layer of complexity to uncertainty management.

Motivated by the critical need for precision in predictions, especially in environments where trade-off between false positives and false negatives carry substantial consequences, we explore the integration of the abstention option with continuous time dynamic graph (CTDGs) learning method. The open problem, therefore, is to incorporate a mechanism that allows CTDGs to abstain from making predictions under high uncertainty, thus reducing the risk of misclassifications in sensitive applications. To address this, we propose a novel framework that integrates the reject option into CTDGs for both link prediction and node classification tasks. This framework aims to enhance model reliability by minimizing the risk associated with uncertain predictions.

We solve this challenge by extending the methodologies of temporal GNNs to incorporate a reject option classification. Our approach involves designing a specialized neural network architecture that dynamically adjusts its confidence threshold based on the evolving graph structure and node interactions. This method allows the model to strategically abstain from making predictions when uncertainty is high, leveraging coverage-based model to optimize the trade-off between accurate prediction and the cost of abstention.

Our main goal is uncertainty management using the reject option. While analyzing the data we find that extreme class imbalance is often present in node classification task for CTDG datasets. As uncertainty amplifies due to class imbalance and noisy boundary between the classes, our approach tries to handle both problems together. 
Previous works have struggled to effectively manage the disproportion between classes, which is particularly problematic in risk-sensitive domains where minority classes may carry significant importance. Our approach handles extreme class imbalance by optimizing the model's training to better represent minority classes, thereby significantly improving performance in these challenging scenarios. This advancement is a pioneering step in dynamic GNNs research, offering a powerful tool for domains where class imbalance profoundly affects utility and reliability of the model. Our main contributions are as follows: 
\begin{enumerate}
\item \textbf{Uncertainty management with reject option:} We integrate a reject option strategy into the framework of CTDG GNNs. This empowers the model to abstain from making predictions when it lacks sufficient confidence, addressing the critical need for risk mitigation in sensitive applications. To achieve this, we propose a coverage-based approach to incorporate the reject option, allowing for the customization of the model's behavior to optimize predictions within a set coverage limit. 

\item \textbf{Managing extreme class imbalance:} Drawing from recent literature for class imbalance mitigation, we address this problem in dynamic graph node classification, an area largely unexplored for extreme imbalance. 

\item \textbf{Comprehensive evaluation:} We conduct extensive experiments on various dynamic graph datasets for link prediction and node classification to demonstrate the effectiveness of our approach. Our results highlight the ability of our models to effectively manage trade-offs between prediction confidence and coverage while significantly improving performance metrics.
\end{enumerate}

\section{Related Work}

Due to the inefficacy of traditional neural networks with data structured as graphs, GNNs\cite{4700287} are the go-to method for dealing with graph data. GNNs address the limitation of learning representations\cite{hamilton2020graph} for nodes and edges. This allows them to capture complex information on how interconnected elements influence each other, enabling them for prediction tasks like node classification (predicting a node's category), link prediction (forecasting the formation of edges), graph classification (classifying graphs based on similarity-dissimilarity), and graph clustering (grouping nodes based on similarity). GNNs employ various message passing schemes (GCN\cite{DBLP:conf/iclr/KipfW17}, GraphSAGE\cite{10.5555/3294771.3294869}, GAT\cite{veličković2018graph} etc.) to aggregate and update node embedding. PyG\cite{Fey/Lenssen/2019} offers a flexible framework based on Pytorch\cite{paszke2017automatic} for GNNs, with variations in layer depth, message passing aggregation that allows to build models for specific graph problems. However, challenges remain in efficiently handling the evolving nature of dynamic graphs, where nodes and edges can change over time.


\noindent\textbf{Temporal Graphs.}
Dynamic graphs primarily operate in two settings\cite{JMLR:v21:19-447}: discrete time\cite{pareja2020evolvegcn} and continuous time\cite{DBLP:conf/kdd/KumarZL19}. In this paper, we focus on CTDGs. There are two types of methods available for handling CTDGs: node-based methods and edge-based methods. Node-based models like TGN\cite{DBLP:journals/corr/abs-2006-10637}, TCL\cite{DBLP:journals/corr/abs-2105-07944}, and DyGFormer\cite{DBLP:conf/nips/YuSLD23} utilize node information, such as temporal neighbors and previous history of nodes to create node embedding. They combine node embeddings to perform predictions for relevant tasks. On the other hand, edge-based methods like GraphMixer\cite{cong2023do} and CAWN\cite{DBLP:journals/corr/abs-2101-05974} directly generate embeddings for the edge of interest. These embeddings are later used for performing tasks like link prediction and node classification in temporal networks. DyGFormer\cite{DBLP:conf/nips/YuSLD23} proposed a transformer-based architecture of a GNN that uses neighbor co-occurrence encoding scheme along with a patching technique to effectively capture long-term temporal dependencies in dynamic graphs. Recent methods DyG-Mamba\cite{li2024dyg}, DyGMamba\cite{ding2024dygmamba} are using node-level and time-level continuous state space model (SSM) to encode node interactions and dynamically select critical temporal information to improve robustness and generalization. Whereas FreeDyG\cite{tian2024freedyg} focus on frequency domain to extract periodic and shifting interaction pattern.
DyGFormer provides a unified library, DyGLib\cite{DBLP:journals/corr/abs-2105-07944} based on PyG\cite{Fey/Lenssen/2019}, with all existing methods implemented in it which supports most of the existing datasets. Similarly, TGB\cite{huang2023temporal} benchmarked over most of the popular temporal graph models for node and edge-level prediction but treated the task as a ranking problem. We use DyGLib extensively for our work to set up experiments and build classification with a rejection module over it. 

\noindent\textbf{Uncertainty Estimation \& Classification With Rejection (CwR).}
CwR here can be classified broadly into two categories: cost-based and coverage-based CwR. Cost-based CwR assumes different cost for rejection. This concept of cost-based method aligns well with cost-sensitive learning\cite{ElYanivWiener2007} where the cost of misclassification can be used to inform the decision to abstain from prediction. Variants of support vector machine (SVM) with reject option are presented in\cite{manwani2015double, shah2019sparse}.\cite{cao2022generalizing} introduces a novel approach to multi-class classification with rejection compatible with arbitrary loss functions, addressing the need for flexibility in adapting to different datasets.
In coverage-based CwR, SelectiveNet\cite{geifman2019selectivenet} introduces a novel approach by integrating the reject option directly within the deep neural network architecture. 
~\cite{NEURIPS2019mcwr, pmlr-v139-charoenphakdee21a} provide calibrated losses for multi-class reject option classification. 
~\cite{cao2022generalizing} propose a general recipe to convert any multi-class loss function to accommodate the reject option, calibrated to loss $l_{0d1}$. They treat rejection as another class. There is a prominent statistical framework in the literature called conformal prediction (CP) that provides valid uncertainty estimates by constructing prediction sets that contain the true label with a predefined confidence level. CF-GNN\cite{huang2024uncertainty} integrates CP with GNNs to generate uncertainty-aware predictions on graph-structured data. In our task for CTDGs, we only handle binary classification. So, there is little scope to apply this method in this setting due to the binary nature of the predictions.

\section{Preliminaries}

This section introduces the concepts and notations used throughout this paper. Our work focuses on extending GNN to temporal graphs, incorporating a classification with a rejection option for enhanced prediction in both link prediction and node classification tasks.\\
\textbf{Deep Learning on Static Graphs.}
It focuses on learning representations for graphs that do not change over time. A static graph is denoted as $\mathcal{G}=(\mathcal{V}, \mathcal{E})$, where $\mathcal{V}$ is the set of nodes and $\mathcal{E} \subseteq \mathcal{V} \times \mathcal{V}$ represents the set of edges connecting these nodes. Each node $u \in \mathcal{V}$ and edge $(u,v) \in \mathcal{E}$ can have associated features $\mathbf{w}_u$ and $\mathbf{e}_{uv}$, respectively. GNNs use features and the structure of the graph to learn a function $f: \mathcal{V} \rightarrow \mathbb{R}^d$. It aggregates information from the local neighborhoods of nodes through a series of message-passing operations that map nodes to a $d$-dimensional embedding space.\\
\textbf{Deep Learning on Dynamic Graphs.}
Dynamic graphs extend the concept of static graphs to accommodate changes over time, representing evolving relationships within the graph. A dynamic graph is represented as a series of graphs $\mathcal{G}_t = (\mathcal{V}_t, \mathcal{E}_t)$ at discrete time steps $t$, or as a continuous stream of interactions $(u,v,t) \in \mathcal{E}_t$, where $u, v \in \mathcal{V}_t$ are nodes, and $t$ represents the time of interaction. The primary goal in deep learning on dynamic graphs is to learn a function $f: \mathcal{V}_t \rightarrow \mathbb{R}^d$ that captures not only the structural features but also its temporal dynamics, facilitating tasks such as temporal link prediction and dynamic node classification.\\
\textbf{Classification with Rejection.}
Rejection classification introduces a decision framework in which a model, given an input $x \in \mathcal{X}$, can choose to abstain from making a prediction if it lacks confidence. Given a prediction model $f: \mathcal{X} \rightarrow \mathcal{Y}$, where $\mathcal{X}$ is the input space and $\mathcal{Y}$ is the output label space, and an abstention function $q: \mathcal{X} \rightarrow \{0, 1\}$. 
{For each sample $x$, the model predicts an abstention score $a(x)$ where $a: \mathcal{X} \in [0, 1]$. We first sort the abstention scores for all the interaction and then select a threshold $\theta$ value that aligns with the desired coverage. For an input $x$, if the abstention score $a(x) \leq \theta$ then $q(x) = 0$ (the model is confident) and the model predicts $f(x)$. In case $a(x) > \theta$ then $q(x) = 1$ and the model abstains the decision for input $x$.} For a given $\theta \in (0, 1)$, the objective is to maximize the probability of correct prediction for the subset of the inputs where $q(x) = 0$:
\begin{equation}
 \max_{f, q} \, P(f(x) = y | q(x) = 0) = \max_{f, q} \, P(f(x) = y | a(x) \leq \theta)
\end{equation}
\textbf{Problem Formulation.}
Given a temporal graph $\mathcal{G}_t$ with tuple $(u, v, t) \in \mathcal{E}_t$ representing an interaction between nodes $u, v \in \mathcal{V}_t$, at time $t \in \mathcal{T}$, our objective is to design a temporal GNN model that addresses the problem of learning effective representations for link prediction and node classification, while incorporating a reject option to manage uncertainty in predictions.\\
\textbf{1. Link Prediction:} Given a temporal graph up to time $(t-1)$, predict the presence or absence of links at time $t$, with the option to abstain from making a prediction for a certain pair of nodes based on a coverage criterion.\\
\textbf{2. Node Classification:} Predict the label of the nodes in a temporal graph using information up to time $(t-1)$, with the option to abstain from the prediction on certain nodes to ensure high confidence in the predictions made.




\section{Method}

\begin{figure*}[t]
\centering
\includegraphics[width=\textwidth, clip]{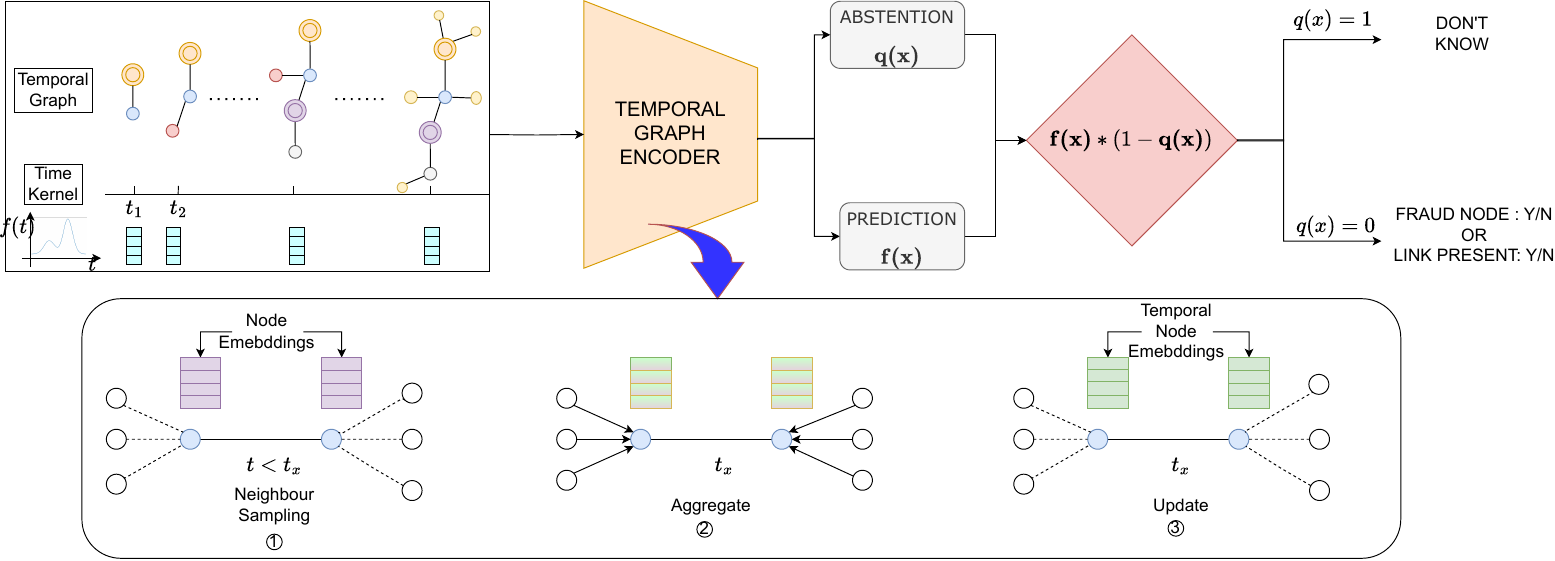}
\caption{The model overview of classification with rejection for task agnostic temporal graphs. Input to the model is a temporal graph with $t$ timestamps, {which is processed through an encoder that samples neighbours from all interactions from previous timestamps $t<t_x$ of these nodes, then aggregates and yields updated temporal node embedding}. These embeddings are passed to a \emph{Prediction} function $f(x)$ to predict downstream tasks such as if a link exists or not for link prediction or if a user is blocked or not for node classification at time $t$. The output embeddings also pass through \emph{Abstention} function, predict a single score and check if 
{$a(x) > \theta$ threshold i.e. $q(x) = 1$}. When it is True, the model abstains, otherwise, it proceeds to make a prediction. The \emph{Auxiliary} head exposes all samples to the model even with high abstention score for a better training process.}
\label{fig:model}
\end{figure*}

In this section, we describe our approach for both link prediction and node classification in dynamic graphs in continuous time using coverage-based method. A neural network architecture designed for dynamic graphs can be seen as a combination of an encoder and a decoder. The encoder serves to transform a dynamic graph into node embeddings, whereas the decoder utilizes these embeddings to make predictions. Our method considers temporal graph networks to incorporate a reject option, enhancing prediction and handling uncertainty effectively. The architecture is illustrated in Fig. \ref{fig:model}.

\subsection{Temporal Graph Encoder}
Our approach is based on the concept of temporal graph networks, which are designed as an encoder-decoder pair for dynamic graph learning. The encoder maps a dynamic graph into the temporal node embeddings $\mathbf{z}_{u}(t)$, capturing the temporal interactions among the nodes. The encoder updates the node embeddings with each interaction, ensuring those reflect the graph's latest state. Few such models are memory-based TGN\cite{DBLP:journals/corr/abs-2006-10637}, MLP-based GraphMixer\cite{cong2023do}, graph transformer-based TCL\cite{DBLP:journals/corr/abs-2105-07944}, random-walk based CAWN\cite{DBLP:journals/corr/abs-2101-05974}, SSM-based DyGMamba\cite{ding2024dygmamba} etc.



\subsection{Coverage Based Dynamic Link Prediction}

Our approach to link prediction uses the concept of abstention prediction\cite{geifman2019selectivenet} to manage the uncertainty inherent in predicting links in a dynamic graph employing an abstention model $(f,q)$ where $f$ is the prediction function and $q$ is the abstention function. The abstention function $q: \mathcal{X} \rightarrow \{0,1\}$, if it takes value one, then the prediction is kept on hold for a given pair of nodes based on a coverage criterion. 
We modify the decoder part of the model to incorporate a reject option through an abstention prediction mechanism. This mechanism evaluates each potential link for its likelihood and the model's confidence in its prediction. By allowing the model to abstain from making predictions when the uncertainty exceeds a threshold $\theta$, we ensure it only predicts links it is confident about. The abstention prediction objective is defined as follows:
\begin{equation}
    \mathcal{L}_{(f,q)}^{\mathcal{E}_t} = \hat{r}(f,q|\mathcal{E}_t) +\lambda \Psi (c-\hat{\phi}(q|\mathcal{E}_t))
\end{equation}
where $\hat{r}(f,q|\mathcal{E}_t)$ is the empirical abstention risk, $\lambda$ is a hyperparameter controlling the importance of the coverage constraint, $\Psi(b)=\max(0,b)^2$ is a quadratic penalty function that helps ensure that the model does not abstain excessively and $c$ is the target coverage rate.
The empirical abstention risk ($\hat{r}(f,q | \mathcal{E}_t)$) and the empirical coverage ($\hat{\phi}(q | \mathcal{E}_t)$) are calculated as follows:
\begin{align}
    \hat{r}(f,q | \mathcal{E}_t) & = \frac{\underset{(u,v,t)\in \mathcal{E}_t}{\sum}\ell(f(\mathbf{z}_{u}(t), \mathbf{z}_{v}(t)), {y}_{\mathcal{E}_{t}}) (1 - q(\mathbf{z}_{u}(t), \mathbf{z}_{v}(t)))}{{\vert \mathcal{E}_t\vert}\;\hat{\phi}(q | \mathcal{E}_t)} \label{eq:r_hat}\\
    \hat{\phi}(q | \mathcal{E}_t) & = \frac{1}{\vert \mathcal{E}_t\vert}\sum_{(u,v,t)\in \mathcal{E}_t}(1 - q(\mathbf{z}_{u}(t), \mathbf{z}_{v}(t))) \label{eq:phi_hat}
\end{align}
where $\ell$ is the \emph{loss function} and $\mathbf{z}_{u}(t)$, $\mathbf{z}_{v}(t)$ are the temporal node embeddings for source and destination node for link prediction respectively and ${y}_{\mathcal{E}_{t}}$ is the binary label of link between two nodes if exists or not. $\hat{\phi}$, the empirical coverage rate of abstention function $q$, calculates the rate at which the edges till time $t$ are selected where ${\vert \mathcal{E}_t\vert}$ is the number of edges. 
The overall training objective combines the abstention loss with an auxiliary loss. For our case, we use the binary cross-entropy loss as an auxiliary loss, denoted as $\mathcal{L}_h^{\mathcal{E}_t}$. The auxiliary loss is computed using the auxiliary prediction head, trained on the same prediction task without considering coverage. This ensures that the model is exposed to all training instances, mitigating the risk of overfitting to a subset of the data. The overall training objective is:
\begin{equation}
\label{ourloss}
    \mathcal{L}^{\mathcal{E}_t} = \alpha \mathcal{L}_{(f,q)}^{\mathcal{E}_t} + (1-\alpha)\mathcal{L}_h^{\mathcal{E}_t}
\end{equation}
where $\alpha$ is a parameter controlling the trade-off between abstention and auxiliary losses.

The temporal graph encoder captures node interaction timing using a time encoder, which is incorporated into the temporal node embeddings. This time information is then passed to both the abstention and prediction functions, allowing the model to decide whether to abstain from making a prediction at a given time $t$.

\subsection{Coverage Based Dynamic Node Classification}
For node classification or user state change prediction\cite{DBLP:conf/kdd/KumarZL19}, we apply a framework similar to link prediction, using the abstention prediction mechanism to introduce a reject option. Node embedding $\mathbf{z}_{u}(t)$, derived from the temporal graph encoder, incorporate information from the node's interaction history and its neighbors to produce a comprehensive representation of its current state in the graph. It serves as input to a classification model that incorporates a coverage-based abstention mechanism. The classifier is designed to accurately predict node label, $y_t \in \mathcal{Y}$ while being able to abstain from predictions when uncertainty exceeds the threshold $\theta$. The abstention prediction objective in this case is as follows:
\begin{equation}
    \mathcal{L}_{(f,q)}^{\mathcal{V}_t} = \hat{r}(f,q|\mathcal{V}_t) +\lambda \Psi (c-\hat{\phi}(q|\mathcal{V}_t))
\end{equation}
Where the empirical abstention risk ($\hat{r}(f,q | \mathcal{V}_t)$) and the empirical coverage ($\hat{\phi}(q | \mathcal{V}_t)$) are calculated as follows:
\begin{align}
    \hat{r}(f,q | \mathcal{V}_t) &= \frac{\sum_{u\in \mathcal{V}_t}\ell(f(\mathbf{z}_{u}(t)), y_t) (1 - q(\mathbf{z}_{u}(t))}{{\vert \mathcal{V}_t\vert}\;\hat{\phi}(q | \mathcal{V}_t)} \label{eq:r_hat_node} \\
    \hat{\phi}(q | \mathcal{V}_t) &= \frac{1}{\vert \mathcal{V}_t\vert}\sum_{u\in \mathcal{V}_t}(1 - q(\mathbf{z}_{u}(t))) \label{eq:phi_hat_node}
\end{align}
{where ${\vert \mathcal{V}_t\vert}$ is the number of nodes.}
The node classification model is trained by minimizing a loss function that balances the precision of the abstention prediction with the desired coverage level which is given by:
\begin{equation}
\label{ourloss_node}
    \mathcal{L}^{\mathcal{V}_t} = \alpha \mathcal{L}_{(f,q)}^{\mathcal{V}_t} + (1-\alpha)\mathcal{L}_h^{\mathcal{V}_t}
\end{equation}
where $\mathcal{L}_h^{\mathcal{V}_t}$ is the auxiliary loss (binary cross-entropy).

\subsection{Handling Extreme Class Imbalance for Dynamic Node Classification}
Dealing with an extreme class imbalance in node classification within CTDGs presents significant challenges. Class imbalance adversely affects model performance, particularly for minority classes, which is crucial for many real-world applications. To address this, we introduce an approach that modifies the training process to better represent minority classes, thereby improving the model's performance. We modify the components of our training objective given in Equation~(\ref{ourloss_node}).\\
\noindent\textbf{Auxiliary Loss for Class Imbalance:} We propose an auxiliary loss function that explicitly accounts for the disproportionate representation of classes by assigning a higher weight to the minority class which is defined as:
\begin{equation}
    \mathcal{L}_h^{\mathcal{V}_t} = \beta  \mathcal{L}_{h_{minor}}^{\mathcal{V}_t} + \mathcal{L}_{h_{major}}^{\mathcal{V}_t}
\end{equation}
where $\mathcal{L}_{h_{minor}}^{\mathcal{V}_t}$ and $\mathcal{L}_{h_{major}}^{\mathcal{V}_t}$ represent the loss for the minority and majority classes, respectively, and $\beta$ is a weighing factor that amplifies the contribution of the minority class to the overall loss. In our experiments, we set hyperparameter $\beta > 1$, highlighting our focus on the minority class, given its significantly lower representation (less than 0.2\% in our dataset). This allows us to tune our model's performance, ensuring that it provides reliable predictions in the case of extreme class imbalance.

\section{Experimental Setup}
In this section we discuss about experimental setup which comprise of baselines, datasets used, evaluation tasks performed, metric used and implementation details.

\subsection{Dataset Used}
We use four distinct datasets for dynamic link prediction and two for dynamic node classification where labels are available. We have chosen popular datasets like Wikipedia\cite{DBLP:conf/kdd/KumarZL19}, Reddit\cite{DBLP:conf/kdd/KumarZL19}, and also included challenging datasets like UN Trade (Economics), Can. Parl. (politics) for diversity. These datasets are publicly available \footnote{\url{https://zenodo.org/record/7213796\#.Y1cO6y8r30o}} by DGB\cite{poursafaei2022towards}. 

\subsection{Baselines Used}
Although our framework is compatible with various temporal graph models supported by DyGLib\cite{DBLP:conf/nips/YuSLD23}, we opted to utilize TGN\cite{DBLP:journals/corr/abs-2006-10637} and GraphMixer\cite{cong2023do} due to their efficiency in terms of time and memory compared to other models, as well as their widespread adoption as baseline. We also included the results for DyGFormer in the appendix as the current . Furthermore, we want to show that our model works on all variations of dynamic graph encoders. 

\noindent\textbf{TGN Encoder \cite{DBLP:journals/corr/abs-2006-10637} :}
The key components of TGN Encoder are:
a) \textbf{Memory module} maintains a compressed representation of each node's historical interactions.
b) \textbf{Message function} computes the impact of new interactions on the state of the node.
c) \textbf{Message aggregator} combines messages from multiple events involving the same node in a batch, improving the update process.
d) \textbf{Memory updater} integrates new interaction messages into the node's memory.
e) \textbf{Embedding module} generates the current embedding of a node using its memory and the memories of its neighbors.

Temporal node embedding is formalized as:
\begin{equation}
\begin{split}
    \mathbf{z}_u(t) &= \sum_{v \in \mathcal{N}^k_u([0, t]) }h(\mathbf{m}_u(t), \mathbf{m}_v(t), \mathbf{e}_{uv}, \mathbf{w}_u(t), \mathbf{w}_v(t))
\end{split}
\end{equation}

where {node $v$ belongs to $k$-hop neighborhood nodes of $u$ between time $0-t$,} $h$ is a learnable function, $\mathbf{m}_u(t)$ and $\mathbf{m}_v(t)$ represent the memory state of source node $u$ and destination node $v$ at time $t$. $\mathbf{e}_{uv}$ denotes edge event features between node $u$ and $v$ and $\mathbf{w}_u(t), \mathbf{w}_v(t)$ are node event features of $u^{th}$ and $v^{th}$ node respectively.

\noindent\textbf{GraphMixer Encoder \cite{cong2023do}:}
GraphMixer presents a simple, straightforward, yet effective approach built upon two primary modules: a link encoder, a node encoder based on multilayer perceptrons (MLPs) without relying on complex GNN architectures. 
a) \textbf{Link-encoder} employs a fixed time-encoding function, $\cos(\boldsymbol{\omega}t)$. It encodes temporal information of links, followed by an MLP-Mixer\cite{DBLP:conf/nips/TolstikhinHKBZU21} to summarize this temporal link information effectively.
b) \textbf{Node-encoder} utilizes neighbor mean-pooling to aggregate and summarize the features of nodes, capturing essential node identity and feature information for subsequent processing steps.

\subsection{Evaluation Tasks and Metrics}
We evaluate the performance of our proposed model in two primary tasks: link prediction and node classification in dynamic graph settings to enhance the practical applicability and trustworthiness of the predictions by following \cite{DBLP:conf/nips/YuSLD23,DBLP:conf/iclr/XuRKKA20,DBLP:journals/corr/abs-2006-10637,DBLP:journals/corr/abs-2101-05974,poursafaei2022towards}.
For the dynamic link prediction task, our objective is to predict the likelihood of a link forming between two nodes at a given time. This task is divided into two settings: transductive and inductive. The \textit{transductive} setting focuses on predicting future links among nodes observed during training, while the \textit{inductive} setting aims at predicting links between the nodes not seen during training. Following established protocols, we utilize a multilayer perceptron (MLP) to process concatenated node embeddings of source and destination and output link probabilities. Evaluation metrics for this task include average precision (AP) and area under the receiver operating characteristic curve (AUC-ROC), with random (rnd), historical (hist) and inductive (ind) negative sampling strategies (NSS)\cite{poursafaei2022towards} employed to improve the evaluation, since the latter two are more challenging. Random NSS randomly selects negative edges from all possible node pairs, risking easy examples and collisions with positive edges. Historical NSS selects negatives from previously observed but currently absent edges, testing prediction of edge re-occurrence. Inductive NSS focuses on edges seen during testing but not training, assessing model generalization to unseen edges.

The dynamic node classification task seeks to identify the label of a node at a specific time. Using the foundation set by the transductive dynamic link prediction model as a pre-trained model on full dataset, we use a separate MLP for label prediction, adhering to the classification with reject option mechanism to allow uncertainty handling. The datasets used for evaluation exclude those without dynamic node labels. The primary metric for this task is AUC-ROC, selected to address label imbalance issues commonly encountered in dynamic graphs\cite{DBLP:conf/iclr/XuRKKA20,DBLP:conf/nips/YuSLD23,DBLP:journals/corr/abs-2006-10637}.

For both tasks, we use a chronological data split strategy, allocating 70\% of the data for training, 15\% for validation and 15\% for testing, to simulate real-world temporal dynamics and ensure a fair evaluation of our model's predictive capabilities.


\subsection{Implementation Details} 
For link prediction task, we optimize both the models by Adam optimizer\cite{DBLP:journals/corr/KingmaB14} for all datasets. For node classification task, we optimize both the models by Adam optimizer for the Wikipedia dataset and SGD for the Reddit dataset. We train the models for 75 epochs and use the early stopping strategy with a patience of 10. We select the model that achieves the best performance in the validation set for testing. We set the batch size to 200, $\lambda$ to 32, $\alpha$ to 0.5 i.e., equal weight to both abstention prediction loss and auxiliary loss for all the methods on all the datasets. We tune the learning rate using cross validation. We search the best value of $\beta$ in the range of [2,100] and we got best results for $\beta=2$ or 5. We have explained more about the choice of hyperparameters in appendix section \ref{Hyperparameters}.
{The selective head of our model gives a proxy of confidence for each sample. Further, we calculate the AUC/AP metrics on only the samples for which the score is higher than the threshold derived using the validation set for each coverage rate by sorting the selection scores for each interaction.} We run each model five times with seeds from 0 to 4 and report the average performance and standard deviation to eliminate deviations. We use NVIDIA GeForce RTX 4090 with 24 GB memory for link prediction task. The GPU device used for the node classification task is NVIDIA GeForce RTX 2080 Ti with 11 GB memory.

\section{Experimental Results}
In this section we report the performance of our proposed continuous time dynamic graph neural network with a reject option. We use four datasets for link prediction and two datasets for node classification task to compare our results with baselines. 

\subsection{Results: Dynamic Link Prediction}

\begin{table*}[!htb]
\renewcommand{\arraystretch}{1.0}
\caption{AP for transductive dynamic link prediction with random, historical, and inductive negative sampling strategies for various coverage rates. The best and second-best results are emphasized by \textbf{bold} and \underline{underlined}.}
\label{tab:transductive_AP}
\centering
\resizebox{\linewidth}{!}{
\begin{tabular}{l|c|cccc|cccr}
\hline
\multirow{2}{*}{NSS}         & \multirow{2}{*}{Coverage} & \multicolumn{4}{c|}{TGN}                                               & \multicolumn{4}{c}{GraphMixer}                                        \\ \cline{3-10}
                             &        (\%)                  & Wikipedia       & Reddit          & UN Trade         & Can. Parl.      & Wikipedia       & Reddit          & UN Trade         & Can. Parl.      \\ \hline
                             \hline
    \multirow{6}{*}{rnd}    & 100                     & 98.56 $\pm$ 0.06 & 98.63 $\pm$ 0.02 & 64.87 $\pm$ 1.93 & 74.14 $\pm$ 1.51 & 97.30 $\pm$ 0.05 & 97.35 $\pm$ 0.02 & 61.68 $\pm$ 0.16 & 79.96 $\pm$ 0.95 \\ \cline{2-10}
                             & 90                     & 99.33 $\pm$ 0.06 & 99.28 $\pm$ 0.07 & 71.67 $\pm$ 1.04 & 75.38 $\pm$ 1.44 & 98.24 $\pm$ 0.07 & 97.66 $\pm$ 0.06 & 68.57 $\pm$ 0.21 & 80.56 $\pm$ 1.71 \\
                             & 80                     & 99.73 $\pm$ 0.08 & 99.54 $\pm$ 0.06 & 76.59 $\pm$ 1.50 & 77.21 $\pm$ 5.87 & 98.89 $\pm$ 0.06 & 98.41 $\pm$ 0.03 & \underline{72.50 $\pm$ 1.24} & 82.44 $\pm$ 1.31 \\
                             & 70                     & 99.85 $\pm$ 0.02 & 99.81 $\pm$ 0.04 & 79.09 $\pm$ 1.66 & 81.57 $\pm$ 7.38 & 99.61 $\pm$ 0.14 & 99.32 $\pm$ 0.08 & 71.64 $\pm$ 8.98 & \underline{95.25 $\pm$ 0.65} \\
                             & 60                     & \textbf{99.88 $\pm$ 0.02} & \underline{99.86 $\pm$ 0.04} & \underline{81.46 $\pm$ 1.81} & \textbf{90.19 $\pm$ 3.50} & \textbf{99.68 $\pm$ 0.11} & \underline{99.52 $\pm$ 0.11} & 69.99 $\pm$ 11.27& \textbf{95.98 $\pm$ 0.57} \\
                             & 50                     & \underline{99.87 $\pm$ 0.05} & \textbf{99.91 $\pm$ 0.03} & \textbf{84.46 $\pm$ 0.95} & \underline{89.46 $\pm$ 6.67}& \underline{99.63 $\pm$ 0.08} & \textbf{99.65 $\pm$ 0.11} & \textbf{80.94 $\pm$ 2.11} & 94.68 $\pm$ 1.25 \\  \bottomrule 

    \multirow{6}{*}{hist}   & 100                      & 87.07 $\pm$ 0.81 & 80.63 $\pm$ 0.30 & 59.44 $\pm$ 2.22 & 70.92 $\pm$ 2.11 & 91.14 $\pm$ 0.16 & 77.50 $\pm$ 0.54 & 57.61 $\pm$ 1.11 & 80.66 $\pm$ 1.37 \\ \cline{2-10}
                             & 90                     & 90.81 $\pm$ 0.59 & 81.68 $\pm$ 1.07 & 70.36 $\pm$ 0.80 & 74.18 $\pm$ 1.94 & 94.25 $\pm$ 0.43 & 78.17 $\pm$ 0.25 & 63.36 $\pm$ 3.08 & 83.10 $\pm$ 1.57 \\
                             & 80                     & 94.46 $\pm$ 0.64 & 83.78 $\pm$ 0.33 & 74.44 $\pm$ 3.55 & 72.82 $\pm$ 6.51 & 96.57 $\pm$ 0.63 & 80.57 $\pm$ 0.33 & 69.35 $\pm$ 3.68 & 84.08 $\pm$ 2.60 \\
                             & 70                     & 95.99 $\pm$ 3.62 & 87.25 $\pm$ 1.76 & 71.94 $\pm$ 8.63 & 76.08 $\pm$ 9.00 & \underline{97.99 $\pm$ 2.64} & 83.61 $\pm$ 1.96 & \underline{75.01 $\pm$ 4.15} & \underline{94.97 $\pm$ 0.55} \\
                             & 60                     & \underline{97.06 $\pm$ 3.84} & \underline{88.66 $\pm$ 1.68} & \underline{76.36 $\pm$ 2.16} & \underline{86.97 $\pm$ 4.56} & \textbf{98.56 $\pm$ 2.81} & \underline{85.32 $\pm$ 1.76} & 69.62 $\pm$ 1.57 & \textbf{95.05 $\pm$ 0.97} \\
                             & 50                     & \textbf{97.82 $\pm$ 4.50} & \textbf{89.27 $\pm$ 2.41} & \textbf{76.65 $\pm$ 2.85} & \textbf{87.34 $\pm$ 3.79} & 97.40 $\pm$ 3.50 & \textbf{85.91 $\pm$ 2.04} & \textbf{81.70 $\pm$ 2.52} & 94.18 $\pm$ 1.33 \\  \bottomrule 

    \multirow{6}{*}{ind}    & 100                    & 86.57 $\pm$ 0.96 & 88.02 $\pm$ 0.35 & 61.70 $\pm$ 2.21 & 68.05 $\pm$ 2.69 & 88.83 $\pm$ 0.13 & 85.21 $\pm$ 0.28 & 60.89 $\pm$ 1.01 & 77.74 $\pm$ 1.42 \\ \cline{2-10}
                             & 90                     & 89.75 $\pm$ 1.03 & 90.17 $\pm$ 0.82 & 72.31 $\pm$ 0.77 & 71.28 $\pm$ 2.90 & 92.11 $\pm$ 0.23 & 85.76 $\pm$ 0.34 & 67.57 $\pm$ 2.52 & 79.98 $\pm$ 2.47 \\
                             & 80                     & 93.19 $\pm$ 0.57 & 92.89 $\pm$ 0.80 & 77.43 $\pm$ 2.94 & 69.06 $\pm$ 9.77 & 95.04 $\pm$ 0.37 & 88.74 $\pm$ 0.43 & 74.08 $\pm$ 1.72 & 79.26 $\pm$ 3.90 \\
                             & 70                     & 95.85 $\pm$ 3.61 & 94.95 $\pm$ 2.27 & 75.75 $\pm$ 6.69 & 72.87 $\pm$ 10.05& \underline{96.45 $\pm$ 3.70} & 91.38 $\pm$ 0.86 & \underline{79.47 $\pm$ 2.22} & 93.47 $\pm$ 0.56 \\
                             & 60                     & \underline{96.76 $\pm$ 4.00} & \textbf{96.57 $\pm$ 2.68} & \textbf{78.65 $\pm$ 1.93} & \underline{84.91 $\pm$ 5.02} & \textbf{97.66 $\pm$ 3.90} & \underline{93.11 $\pm$ 1.45} & 77.98 $\pm$ 1.54 & \underline{94.11 $\pm$ 0.84} \\
                             & 50                     & \textbf{97.75 $\pm$ 4.63} & \underline{95.93 $\pm$ 3.33} & \underline{78.30 $\pm$ 2.15} & \textbf{85.12 $\pm$ 6.52} & 94.79 $\pm$ 6.29 & \textbf{94.35 $\pm$ 1.60} & \textbf{85.47 $\pm$ 2.57} & \textbf{94.28 $\pm$ 1.07} \\  \bottomrule 

\end{tabular}}
\end{table*}


\begin{table*}[!htb]
\caption{AUC for inductive dynamic link prediction with random, historical, and inductive negative sampling strategies for various coverage rates. The best and second-best results are emphasized by \textbf{bold} and \underline{underlined}.}
\label{tab:Inductive_AUC}
\centering
\resizebox{\linewidth}{!}{
\begin{tabular}{c|c|cccc|cccc}
\hline
\multirow{2}{*}{NSS}         & \multirow{2}{*}{Coverage} & \multicolumn{4}{c|}{TGN}                                               & \multicolumn{4}{c}{GraphMixer}                                        \\ \cline{3-10}
                             &               (\%)           & Wikipedia       & Reddit          & UN Trade         & Can. Parl.      & Wikipedia       & Reddit          & UN Trade         & Can. Parl.      \\ \hline
                             \hline
\multirow{6}{*}{rnd}         & 100                     & 97.87 $\pm$ 0.08 & 97.17 $\pm$ 0.04 & 57.82 $\pm$ 2.42 & 60.04 $\pm$ 0.90 & 96.52 $\pm$ 0.09 & 94.86 $\pm$ 0.01 & 63.74 $\pm$ 0.24 & 64.77 $\pm$ 0.79 \\ \cline{2-10} 
                             & 90                      & 99.08 $\pm$ 0.09 & 98.21 $\pm$ 0.13 & 59.93 $\pm$ 1.36 & 60.86 $\pm$ 3.29 & 97.88 $\pm$ 0.14 & 95.71 $\pm$ 0.14 & 66.12 $\pm$ 0.25 & 64.53 $\pm$ 0.79 \\ 
                             & 80                      & 99.62 $\pm$ 0.09 & 98.81 $\pm$ 0.12 & \underline{62.72 $\pm$ 2.59} & 61.67 $\pm$ 2.03 & 98.68 $\pm$ 0.09 & 96.76 $\pm$ 0.13 & 67.80 $\pm$ 0.84 & 64.67 $\pm$ 0.76 \\ 
                             & 70                      & 99.83 $\pm$ 0.06 & 99.54 $\pm$ 0.16 & \textbf{62.75 $\pm$ 1.29} & 63.15 $\pm$ 4.24 & \underline{99.56 $\pm$ 0.29} & 98.63 $\pm$ 0.15 & 68.44 $\pm$ 3.32 & \underline{68.44 $\pm$ 2.62} \\ 
                             & 60                      & \textbf{99.88 $\pm$ 0.03} & \textbf{99.73 $\pm$ 0.13} & 59.32 $\pm$ 4.62 & \underline{66.66 $\pm$ 5.34} & \textbf{99.66 $\pm$ 0.21} & \underline{98.94 $\pm$ 0.24} & 67.51 $\pm$ 5.34 & \textbf{72.40 $\pm$ 4.43} \\ 
                             & 50                      & \underline{99.87 $\pm$ 0.03} & \underline{99.73 $\pm$ 0.14} & 59.95 $\pm$ 3.45 & \textbf{69.72 $\pm$ 3.09} & 99.47 $\pm$ 0.18 & \textbf{99.21 $\pm$ 0.13} & \textbf{73.38 $\pm$ 1.45} & \underline{71.77 $\pm$ 2.01} \\ \bottomrule

\multirow{6}{*}{hist}        & 100                     & 76.83 $\pm$ 0.89 & 62.40 $\pm$ 0.19 & 55.41 $\pm$ 1.74 & 60.16 $\pm$ 0.78 & 83.23 $\pm$ 0.22 & 62.64 $\pm$ 0.19 & 62.21 $\pm$ 0.99 & 64.75 $\pm$ 0.62 \\ \cline{2-10} 
                             & 90                      & 82.01 $\pm$ 0.74 & 64.90 $\pm$ 0.67 & 57.69 $\pm$ 2.43 & 60.49 $\pm$ 2.95 & 88.77 $\pm$ 0.48 & 65.00 $\pm$ 0.28 & 64.05 $\pm$ 1.60 & 64.98 $\pm$ 0.61 \\ 
                             & 80                      & 87.59 $\pm$ 0.53 & 67.18 $\pm$ 0.94 & \textbf{60.71 $\pm$ 2.89} & 60.83 $\pm$ 2.05 & 93.18 $\pm$ 0.84 & 67.73 $\pm$ 0.32 & 66.46 $\pm$ 1.71 & 65.26 $\pm$ 0.69 \\ 
                             & 70                      & 91.34 $\pm$ 8.08 & 69.58 $\pm$ 1.35 & \underline{60.06 $\pm$ 3.28} & 64.16 $\pm$ 4.80 & \underline{95.10 $\pm$ 6.44} & 69.90 $\pm$ 1.65 & 71.18 $\pm$ 1.48 & 70.21 $\pm$ 2.41 \\ 
                             & 60                      & \underline{93.64 $\pm$ 8.72} & \textbf{71.61 $\pm$ 2.34} & 58.78 $\pm$ 3.76 & \underline{67.21 $\pm$ 5.40} & \textbf{96.34 $\pm$ 6.92} & \underline{71.95 $\pm$ 2.24} & \underline{71.28 $\pm$ 3.03} & \textbf{73.57 $\pm$ 4.14} \\ 
                             & 50                      & \textbf{95.31 $\pm$ 9.71} & \underline{70.86 $\pm$ 2.72} & 59.30 $\pm$ 3.32 & \textbf{69.30 $\pm$ 3.85} & 92.26 $\pm$ 10.32& \textbf{73.85 $\pm$ 3.12} & \textbf{76.71 $\pm$ 2.41} & \underline{73.32 $\pm$ 1.62} \\ \bottomrule

\multirow{6}{*}{ind}         & 100                     & 76.82 $\pm$ 0.90 & 62.40 $\pm$ 0.19 & 55.37 $\pm$ 1.73 & 60.14 $\pm$ 0.78 & 83.23 $\pm$ 0.23 & 62.63 $\pm$ 0.19 & 62.18 $\pm$ 0.99 & 64.74 $\pm$ 0.62 \\ \cline{2-10} 
                             & 90                      & 82.01 $\pm$ 0.74 & 64.90 $\pm$ 0.67 & 57.66 $\pm$ 2.43 & 60.46 $\pm$ 2.97 & 88.77 $\pm$ 0.48 & 65.01 $\pm$ 0.28 & 64.02 $\pm$ 1.60 & 64.99 $\pm$ 0.59 \\ 
                             & 80                      & 87.59 $\pm$ 0.54 & 67.18 $\pm$ 0.95 & \textbf{60.69 $\pm$ 2.88} & 60.80 $\pm$ 2.10 & 93.17 $\pm$ 0.84 & 67.75 $\pm$ 0.33 & 66.45 $\pm$ 1.79 & 65.30 $\pm$ 0.74 \\ 
                             & 70                      & 91.35 $\pm$ 8.08 & 69.57 $\pm$ 1.36 & \underline{60.03 $\pm$ 3.26} & 64.22 $\pm$ 4.85 & \underline{95.09 $\pm$ 6.45} & 69.92 $\pm$ 1.67 & 71.21 $\pm$ 1.46 & 70.21 $\pm$ 2.39 \\ 
                             & 60                      & \underline{93.64 $\pm$ 8.73} & \textbf{71.62 $\pm$ 2.34} & 58.74 $\pm$ 3.76 & \underline{67.16 $\pm$ 5.41} & \textbf{96.34 $\pm$ 6.93} & \underline{71.97 $\pm$ 2.26} & \underline{71.33 $\pm$ 3.09} & \textbf{73.55 $\pm$ 4.13} \\ 
                             & 50                      & \textbf{95.32 $\pm$ 9.73} & \underline{70.86 $\pm$ 2.72} & 59.28 $\pm$ 3.34 & \textbf{69.24 $\pm$ 3.87} & 92.24 $\pm$ 10.32& \textbf{73.88 $\pm$ 3.15} & \textbf{76.64 $\pm$ 2.43} & \underline{73.32 $\pm$ 1.60} \\ \bottomrule 
\end{tabular}}
\end{table*}


For dynamic link prediction, we evaluate our model on the Wikipedia, Reddit UnTrade and Can. Parl. datasets. Table \ref{tab:transductive_AP} summarizes the results for Average Precision (AP) metric for transductive dynamic link prediction with three negative sampling strategies (NSS). All values are reported in percentage. The best and second-best performing results are emphasized by \textbf{bold} and \underline{underlined} fonts. 
{Table \ref{tab:Inductive_AUC} presents the results for inductive link prediction for AUC.} 
Our models demonstrate strong performance improvements with the introduction of rejection capabilities.

In the first row of all the tables, scores are reported by running both TGN and GraphMixer models from DyGLib \cite{DBLP:conf/nips/YuSLD23} without the reject option (100\% coverage), which sets our baseline. The scores for all three NSS, both the models and all the datasets progressively improve as rejecting 10\% of samples below the confidence threshold in each subsequent test up to a 50\% coverage level. Our results show a consistent improvement in AP and AUC metrics as the coverage decreases and the rejection rate increases.
{For example, in Table \ref{tab:Inductive_AUC} in Random NSS setting, we get $99.73\%$ AUC (fifth row) in $60\%$ coverage w.r.t. $97.17\%$ AUC (first row) in $100\%$ coverage for TGN model on Reddit dataset marking almost $2.56\%$ improvement in AUC.}
This indicates the model's ability to effectively recognize and reject uncertain samples, thereby focusing on high-confidence predictions.

Challenging datasets such as the Canadian Parliament (Can. Parl.) and UN Trade have been specifically highlighted due to their lower baseline scores, which presents a significant opportunity for improvement. With our model, the AP/AUC score notably increases, emphasizing the model's efficiency in managing harder prediction environments. For example, in Table \ref{tab:transductive_AP} in Random NSS setting, we achieve $90.19\%$ AP (fifth row) in $60\%$ coverage w.r.t. $74.14\%$ AP (first row) in $100\%$ coverage for the TGN model in Can. Parl. dataset marking almost $16.05\%$ improvement in AP. This suggests that our approach is particularly valuable in scenarios where datasets are difficult or highly noisy. One observation is that AP/AUC increases to a certain point (peak) as coverage decreases. It starts decreasing as coverage goes on decreasing. 
We also provide results with DyGFormer encoder that show similar patterns which are moved to appendix \ref{Additional_Results} for brevity.

\subsection{Results: Dynamic Node Classification}

For dynamic node classification, we evaluate our model on the Wikipedia and Reddit datasets, as shown in Table \ref{tab:Node_AUC}. The initial results with 100\% coverage use DyGLib, which establishes a benchmark. As we implement the reject option, progressively rejecting 10\% of samples in each subsequent test until we reach 60\% coverage, we observe a significant improvement in AUC scores (where $\beta=1$). For example in Table \ref{tab:Node_AUC} we achieve $89.78\%$ AUC (ninth row) in $60\%$ coverage w.r.t $86.23\%$ AUC (first row) in $100\%$ coverage for TGN model on Wikipedia marking almost $3.55\%$ improvement in AUC.

This increase in AUC with decreasing coverage indicates that the model is effectively prioritizing more reliable predictions, as less certain predictions are rejected. Furthermore, due to the extreme imbalance in class distribution (minority class below 0.2\%), we provide higher weightage to minority class in the auxiliary loss, which led to further improvements in AUC (where $\beta>1$). For example, in Table \ref{tab:Node_AUC}, we get $69.58\%$ AUC (tenth row) with $\beta=5$ w.r.t $66.03\%$ AUC (ninth row) in $60\%$ coverage for TGN model on Reddit dataset. This marks almost $3.55\%$ improvement in AUC while addressing the class imbalances. This strategy improves the efficacy of our approach in not only managing the reject option but also addressing extreme class imbalances, which are particularly challenging in dynamic environments. Results with DyGFormer\cite{DBLP:conf/nips/YuSLD23} encoder are shown in appendix \ref{Additional_Results} due to space constraints.

\begin{table}[!htb]
\caption{AUC for coverage-based dynamic node classification for various coverage rates with and without handling class imbalance. The best and second-best performing results are emphasized by \textbf{bold} and \underline{underlined} fonts. $^*$ denotes $\beta$ is 2 otherwise 5 for all other cases where $\beta > 1$. }
\vspace{2mm}
\label{tab:Node_AUC}
\centering
\resizebox{0.95\linewidth}{!}{
\begin{tabular}{c|c|c|c|c|c}
\hline
\multirow{2}{*}{Cover-} & \multirow{2}{*}{$\beta$}       & \multicolumn{2}{c|}{TGN}         & \multicolumn{2}{c}{GraphMixer}                                         \\ \cline{3-6}
age(\%) &  & Wikipedia         & Reddit &  Wikipedia        & Reddit \\ \hline \hline
100   & $=1$ & 86.23 $\pm$ 3.30 & 63.08 $\pm$ 1.48 & 86.19 $\pm$ 1.10    & 64.81 $\pm$ 2.08 \\ \hline
\multirow{2}{*}{90}    & $=1$ & 88.00 $\pm$ 2.00 & 64.81 $\pm$ 1.54 & 87.27 $\pm$ 1.72    & 65.26 $\pm$ 2.07 \\
    & $>1$ & {88.44 $\pm$ 2.33}$^*$ & 65.12 $\pm$ 1.53 & {87.61 $\pm$ 1.06}$^*$ & 66.20 $\pm$ 1.88\\ \hline
\multirow{2}{*}{80}    & $=1$ & 88.78 $\pm$ 3.46 & 65.63 $\pm$ 1.31 & 87.93 $\pm$ 1.53 & 66.08 $\pm$ 2.99 \\
    & $>1$ & \underline{90.02 $\pm$ 2.29}$^*$ & 66.31 $\pm$ 3.26 & \underline{88.05 $\pm$ 1.72} & 67.00 $\pm$ 2.40\\ \bottomrule
\multirow{2}{*}{70}    & $=1$ & 89.48 $\pm$ 2.53 & 66.10 $\pm$ 2.34 & 87.79 $\pm$ 1.37 & 67.42 $\pm$ 3.27 \\
    & $>1$ &\textbf{ 90.64 $\pm$ 3.37} & \underline{67.90 $\pm$ 1.81} & \textbf{88.38 $\pm$ 0.99}$^*$ & \textbf{68.23 $\pm$ 2.96} \\ \bottomrule
\multirow{2}{*}{60}    & $=1$ & 89.78 $\pm$ 2.76 & 66.03 $\pm$ 3.45 & 85.97 $\pm$ 2.62 & 67.23 $\pm$ 3.36 \\
    & $>1$ & 89.86 $\pm$ 2.52 & \textbf{69.58 $\pm$ 2.96} & {84.83 $\pm$ 2.30}$^*$ & \underline{67.87 $\pm$ 4.01} \\ \bottomrule
\end{tabular}}
\end{table}

In real-world scenarios we can't reject 40-50\% prediction. But as we can see, with decrease in coverage, performance keeps on increasing. Note that the model rejects examples it does not predict very confidently. We have to find a balance between the rate of abstention and the model confidence based on the application domain, number of data etc.
Overall, the experiments validate our hypothesis that integrating a reject option within CTDGs significantly enhances the reliability and accuracy of predictions in dynamic graphs. The clear improvement in performance metrics across different datasets, coverage settings, and tasks confirms the practical utility of our model in risk-sensitive applications, where precision and reliability are crucial.

\section{Conclusion}
The experimental findings convincingly demonstrate the efficacy of our proposed CTDG model with a reject option for both link prediction and node classification tasks. The model strategically abstains from making predictions when encountering high uncertainty, leading to a significant improvement in AUC/AP scores. This establishes the model's proficiency in managing the trade-off between prediction confidence and coverage. 
In future, we can explore cost-based or other abstention methods to enhance the model's applicability in different contexts. Another future direction could involve demonstrating the model's performance in real-world applications, showcasing its practical relevance. Considering our method for another kind of graph model, such as discrete time temporal graphs, can also be explored. Overall, our work presents a compelling solution for dynamic graph applications that demand high precision and reliability, particularly in risk-sensitive domains. The model's ability to effectively manage uncertainty and class imbalance makes it a valuable tool for various real-world applications.

\bibliographystyle{plain}  
\bibliography{cite}

\newpage
\appendix
\section{Datasets}
We use four publicly available datasets in our experiments detailed below. The dataset statistics are shown in Table \ref{tab:data_statistics}.
\begin{itemize}
\item\textbf{Reddit \cite{DBLP:conf/kdd/KumarZL19}:} A month's worth of posts from users across the 984 most active subreddits, resulting in a total of 672,447 interactions from the 10,000 most active users. Each post's content is transformed into a feature vector using LIWC \cite{pennebaker2001linguistic}. This dataset also includes dynamic labels indicating whether users are banned from posting, with 366 true banned labels, marking a proportion of 0.05\% of the interactions.
\item\textbf{Wikipedia \cite{DBLP:conf/kdd/KumarZL19}:} Comprising one month of Wikipedia edits, this dataset captures interactions from users who have made edits on the 1,000 most edited pages, totaling 157,474 interactions from 8,227 users. Each interaction is represented as a 172-dimensional LIWC feature vector. Additionally, dynamic labels are provided to indicate temporary bans of users, with 217 positive banned labels among the interactions, equating to 0.14\%.
\item\textbf{Canadian Parliament (Can. Parl.) \cite{huang2020laplacian}:} This dynamic political network dataset records interactions between Members of Parliament (MPs) in Canada from 2006 to 2019, where nodes represent MPs, and links indicate mutual "yes" votes on bills. The weight of a link reflects the frequency of mutual affirmative votes within a year.
\item\textbf{UN Trade \cite{poursafaei2022towards}:} Spanning over 30 years, this dataset documents food and agriculture trade between 181 nations. The links between nations are weighted by the normalized value of agriculture imports or exports, showing the intensity of trade relationships.
\end{itemize}

\begin{table}[!htbp]
\centering
\caption{Datasets. \#N\&L Feat. denotes node and link feature dimension.}
\vspace{1mm}
\label{tab:data_statistics}
\resizebox{0.75\linewidth}{!}
{
\begin{tabular}{l|cccccccr}
\hline
Datasets    & \#Nodes & \#Links  & \#N\&L Feat.      & Bipartite & Time Granularity & Label  \\ \hline
Wikipedia   & 9,227  & 157,474   & -- \& 172         & True     &  Unix            & 2  \\
Reddit      & 10,984 & 672,447   & -- \& 172         & True     &  Unix            & 2  \\
Can. Parl.  & 734    & 74,478    & -- \& 1           & False    &  Years           & -- \\
UN Trade    & 255    & 507,497   & -- \& 1           & False    &  Years           & -- \\ \hline
\end{tabular}}
\end{table}

\section{Hyperparameters}
\label{Hyperparameters}
All three encoders use 100 dimensional time encoding, and 172 dimensional output representation. TGN uses a 172-dimensional node memory and uses 2 graph attention heads like DyGFormer , with memory updates managed by Gated Recurrent Units (GRU).
GraphMixer does not use node memory but instead have 2 MLP-Mixer layers and operates with a time gap of 2000. DyGFormer introduces a 50-dimensional neighbor co-occurrence encoding alongside the time encoding and employs 2 Transformer layers to process the input.

We perform the grid search to find the best settings of some critical hyperparameters. However, this step is conducted during training. Once the model is trained, the number of covered examples on the test dataset will vary significantly as $\theta$ changes because the model has not seen the test data during training. Nevertheless, with the selection scores for each interaction available, we sorted them and selected a $\theta$ value that aligns with the desired coverage. The $\theta$ values range from 0 to 1. As the coverage decreases to 90\%, 80\%, and so on down to 50\%, the uncertainty threshold $\theta$ also decreases accordingly. 

We experimented with wide range of $\beta$ values as hyperparameter and found when $\beta$ value is less than 10 it produces better result. We're attaching the following table for TGN on Wikipedia dataset to show the variation AUC w.r.t. different values of beta.
Below is the comparison between $\beta$ vs AUC shown in Table \ref{tab:beta-effect-table}.

\begin{table}[!htb]
\centering
\caption{Effect of $\beta$ on AUC at coverage 90\%, 80\% and 70\% respectively for TGN encoder on Wikipedia dataset}
\vspace{2mm}
\label{tab:beta-effect-table}
\resizebox{0.45\linewidth}{!}{%
\begin{tabular}{lccc}
\hline
$\beta$ & AUC@90 & AUC@80 & AUC@70 \\ \hline
1    & 88     & 88.78  & 89.48  \\
2    & 88.44  & 90.02  & 90.28  \\
5    & 87.96  & 89.7   & 90.64  \\
10   & 87.56  & 89.14  & 89.98  \\
25   & 87.67  & 88.29  & 89.42  \\
100  & 86.84  & 88.32  & 88.74  \\ \hline
\end{tabular}%
}
\end{table}
\section{Compute Statistics}
We have compared compute statistics for different temporal graph encoders for Can. Parl. dataset in table \ref{tab:Compute-Statistics-table} on link prediciton task. We train maximum 100 epochs for training. Based on the following table we have chosen to use TGN and GraphMixer as our baseline.

\begin{table*}[!ht]
\centering
\caption{Compute Statistics for Can. Parl. dataset on link prediction}
\label{tab:Compute-Statistics-table}
\resizebox{0.5\linewidth}{!}{%
\begin{tabular}{lccc}
\hline
\multicolumn{1}{c}{TG Encoder} &
  \begin{tabular}[c]{@{}c@{}}Train Time \\ (Sec/epoch)\end{tabular} &
  \begin{tabular}[c]{@{}c@{}}Test Time\\  (Sec)\end{tabular} &
  \begin{tabular}[c]{@{}c@{}}GPU Usage\\  (MB)\end{tabular} \\ \midrule
TGN        & 9   & 1  & 668   \\
TGAT       & 75  & 22 & 2800  \\
GraphMixer & 27  & 6  & 1708  \\
CAWN       & 161 & 36 & 12114 \\
DyGFormer  & 91  & 20 & 18326 \\
TCL        & 15  & 1  & 1616  \\
Jodie      & 4   & 1  & 640   \\ \bottomrule
\end{tabular}%
}
\end{table*}
\section{Additional Results}
\label{Additional_Results}
Tables \ref{tab:Inductive_AP} and  \ref{tab:transductive_AUC} present the results for inductive and transductive link prediction for AP and AUC-ROC, respectively. We have added link prediction results for DyGFormer\cite{DBLP:conf/nips/YuSLD23} on all four datasets in Table \ref{tab:Trans_AP_AUC} for transductive setting and Table \ref{tab:Ind_AP_AUC} for inductive setting. We have also added node classification results for DyGFormer in Table \ref{tab:dygformer_node_clf}.
\begin{table*}[!htb]
\caption{AP for inductive dynamic link prediction with random, historical, and inductive negative sampling strategies for various coverage rates. The best and second-best performing results are emphasized by \textbf{bold} and \underline{underlined} fonts.}
\label{tab:Inductive_AP}
\centering
\resizebox{\linewidth}{!}{
\begin{tabular}{c|c|cccc|cccc}
\hline
\multirow{2}{*}{NSS}         & \multirow{2}{*}{Coverage} & \multicolumn{4}{c|}{TGN}                                               & \multicolumn{4}{c}{GraphMixer}                                        \\ \cline{3-10}
                             &       (\%)                   & Wikipedia       & Reddit          & UN Trade         & Can. Parl.      & Wikipedia       & Reddit          & UN Trade         & Can. Parl.      \\ \hline
                             \hline
\multirow{6}{*}{rnd}         & 100                     & 97.98 $\pm$ 0.08 & 97.34 $\pm$ 0.05 & 56.55 $\pm$ 2.24 & 57.21 $\pm$ 1.22 & 96.82 $\pm$ 0.06 & 95.13 $\pm$ 0.01 & 61.83 $\pm$ 0.09 & 59.38 $\pm$ 1.05 \\ \cline{2-10} 
                             & 90                      & 99.08 $\pm$ 0.06 & 98.23 $\pm$ 0.14 & 63.23 $\pm$ 1.72 & 57.08 $\pm$ 2.88 & 98.03 $\pm$ 0.12 & 95.76 $\pm$ 0.14 & 68.16 $\pm$ 0.19 & 58.96 $\pm$ 0.85 \\ 
                             & 80                      & 99.57 $\pm$ 0.07 & 98.79 $\pm$ 0.10 & \textbf{66.04 $\pm$ 3.86} & 57.77 $\pm$ 4.64 & 98.76 $\pm$ 0.08 & 96.76 $\pm$ 0.12 & \underline{70.97 $\pm$ 1.19} & 58.83 $\pm$ 0.99 \\ 
                             & 70                      & 99.77 $\pm$ 0.05 & 99.45 $\pm$ 0.09 & \underline{65.51 $\pm$ 2.31} & 64.57 $\pm$ 8.02 & \underline{99.49 $\pm$ 0.14} & 98.39 $\pm$ 0.23 & 70.54 $\pm$ 7.44 & 70.65 $\pm$ 3.30 \\ 
                             & 60                      & \textbf{99.82 $\pm$ 0.06} & \underline{99.68 $\pm$ 0.04} & 61.23 $\pm$ 5.83 & \underline{69.82 $\pm$ 6.54} & \textbf{99.59 $\pm$ 0.12} & \underline{98.75 $\pm$ 0.25} & 68.45 $\pm$ 9.95 & \underline{76.29 $\pm$ 4.96} \\ 
                             & 50                      & \underline{99.80 $\pm$ 0.08} & \textbf{99.73 $\pm$ 0.04} & 60.69 $\pm$ 5.56 & \textbf{71.55 $\pm$ 7.14} & 99.46 $\pm$ 0.09 & \textbf{99.03 $\pm$ 0.29} & \textbf{78.39 $\pm$ 0.90} & \textbf{76.44 $\pm$ 2.39} \\ \bottomrule

\multirow{6}{*}{hist}        & 100                     & 82.12 $\pm$ 1.37 & 61.59 $\pm$ 0.48 & 52.96 $\pm$ 1.27 & 56.75 $\pm$ 1.29 & 87.63 $\pm$ 0.28 & 61.73 $\pm$ 0.36 & 56.41 $\pm$ 0.95 & 58.64 $\pm$ 0.77 \\ \cline{2-10} 
                             & 90                      & 84.97 $\pm$ 1.13 & 61.84 $\pm$ 0.66 & 59.12 $\pm$ 3.88 & 55.95 $\pm$ 2.40 & 91.30 $\pm$ 0.47 & 62.42 $\pm$ 0.29 & 60.48 $\pm$ 2.44 & 58.70 $\pm$ 0.90 \\ 
                             & 80                      & 88.85 $\pm$ 0.83 & 63.02 $\pm$ 0.89 & \textbf{61.20 $\pm$ 4.00} & 56.47 $\pm$ 4.78 & 94.22 $\pm$ 1.03 & 63.66 $\pm$ 0.27 & 65.74 $\pm$ 4.66 & 58.85 $\pm$ 1.53 \\ 
                             & 70                      & 92.42 $\pm$ 4.94 & 66.99 $\pm$ 4.34 & 58.80 $\pm$ 5.99 & 65.27 $\pm$ 9.09 & \underline{96.01 $\pm$ 3.95} & 66.98 $\pm$ 3.44 & \underline{74.08 $\pm$ 3.63} & 73.00 $\pm$ 2.67 \\ 
                             & 60                      & \underline{94.81 $\pm$ 5.67} & \underline{67.96 $\pm$ 3.89} & 58.90 $\pm$ 4.34 & \underline{70.98 $\pm$ 6.00} & \textbf{97.33 $\pm$ 4.50} & \underline{68.73 $\pm$ 3.33} & 70.64 $\pm$ 1.50 & \underline{77.83 $\pm$ 4.37} \\ 
                             & 50                      & \textbf{96.63 $\pm$ 6.68} & \textbf{68.83 $\pm$ 6.00} & \underline{59.02 $\pm$ 5.16} & \textbf{71.27 $\pm$ 8.94} & 94.79 $\pm$ 6.78 & \textbf{69.43 $\pm$ 3.38} & \textbf{82.22 $\pm$ 1.63} & \textbf{78.41 $\pm$ 1.74} \\ \bottomrule 
 
\multirow{6}{*}{ind}         & 100                     & 82.12 $\pm$ 1.38 & 61.59 $\pm$ 0.48 & 52.90 $\pm$ 1.25 & 56.78 $\pm$ 1.31 & 87.63 $\pm$ 0.28 & 61.74 $\pm$ 0.35 & 56.35 $\pm$ 0.93 & 58.63 $\pm$ 0.79 \\ \cline{2-10} 
                             & 90                      & 84.98 $\pm$ 1.13 & 61.85 $\pm$ 0.66 & 59.06 $\pm$ 3.89 & 55.97 $\pm$ 2.41 & 91.30 $\pm$ 0.47 & 62.44 $\pm$ 0.28 & 60.31 $\pm$ 2.54 & 58.72 $\pm$ 0.92 \\ 
                             & 80                      & 88.86 $\pm$ 0.82 & 63.01 $\pm$ 0.90 & \textbf{61.16 $\pm$ 3.98} & 56.53 $\pm$ 4.87 & 94.22 $\pm$ 1.03 & 63.69 $\pm$ 0.26 & 65.64 $\pm$ 4.79 & 58.89 $\pm$ 1.61 \\ 
                             & 70                      & 92.45 $\pm$ 4.95 & 66.98 $\pm$ 4.35 & 58.78 $\pm$ 5.95 & 65.39 $\pm$ 9.15 & \underline{96.01 $\pm$ 3.95} & 67.00 $\pm$ 3.44 & \underline{74.13 $\pm$ 3.50} & 73.01 $\pm$ 2.65 \\ 
                             & 60                      & \underline{94.80 $\pm$ 5.68} & \underline{67.98 $\pm$ 3.88} & 58.83 $\pm$ 4.36 & \underline{70.90 $\pm$ 6.02} & \textbf{97.33 $\pm$ 4.51} & \underline{68.76 $\pm$ 3.31} & 70.65 $\pm$ 1.47 & \underline{77.81 $\pm$ 4.34} \\ 
                             & 50                      & \textbf{96.64 $\pm$ 6.69} & \textbf{68.84 $\pm$ 5.99} & \underline{58.98 $\pm$ 5.22} & \textbf{71.21 $\pm$ 8.90} & 94.79 $\pm$ 6.78 & \textbf{69.48 $\pm$ 3.35} & \textbf{82.16 $\pm$ 1.72} & \textbf{78.45 $\pm$ 1.73} \\ \bottomrule 
 
\end{tabular}}
\end{table*}
\begin{table*}[!htb]
\caption{AUC for transductive dynamic link prediction with random, historical, and inductive negative sampling strategies for various coverage rates. The best and second-best performing results are emphasized by \textbf{bold} and \underline{underlined} fonts.}
\label{tab:transductive_AUC}
\centering
\resizebox{\linewidth}{!}{
\begin{tabular}{c|c|cccc|cccc}
\hline
\multirow{2}{*}{NSS}         & \multirow{2}{*}{Coverage} & \multicolumn{4}{c|}{TGN}                                               & \multicolumn{4}{c}{GraphMixer}                                        \\ \cline{3-10}
                             &           (\%)               & Wikipedia       & Reddit          & UN Trade         & Can. Parl.      & Wikipedia       & Reddit          & UN Trade         & Can. Parl.      \\ \hline
                             \hline
\multirow{6}{*}{rnd}        & 100                     & 98.48 $\pm$ 0.07 & 98.59 $\pm$ 0.02 & 67.74 $\pm$ 1.66 & 77.94 $\pm$ 1.55 & 97.01 $\pm$ 0.07 & 97.20 $\pm$ 0.02 & 64.89 $\pm$ 0.21 & 84.64 $\pm$ 0.57 \\ \cline{2-10} 
                             & 90                      & 99.32 $\pm$ 0.05 & 99.31 $\pm$ 0.06 & 70.24 $\pm$ 0.92 & 80.17 $\pm$ 1.44 & 98.04 $\pm$ 0.08 & 97.63 $\pm$ 0.07 & 68.01 $\pm$ 0.21 & 85.39 $\pm$ 1.04 \\ 
                             & 80                      & 99.75 $\pm$ 0.09 & 99.57 $\pm$ 0.06 & 74.50 $\pm$ 1.21 & 82.32 $\pm$ 2.21 & 98.72 $\pm$ 0.07 & 98.40 $\pm$ 0.03 & 70.52 $\pm$ 0.70 & 86.77 $\pm$ 0.94 \\ 
                             & 70                      & 99.88 $\pm$ 0.06 & 99.85 $\pm$ 0.02 & 76.71 $\pm$ 1.64 & 82.05 $\pm$ 5.10 & \underline{99.63 $\pm$ 0.28} & 99.41 $\pm$ 0.06 & \underline{71.36 $\pm$ 4.07} & \underline{93.05 $\pm$ 1.10} \\ 
                             & 60                      & \underline{99.90 $\pm$ 0.04} & \underline{99.90 $\pm$ 0.02} & \underline{79.25 $\pm$ 1.86} & \underline{88.03 $\pm$ 3.69} & \textbf{99.72 $\pm$ 0.22} & \underline{99.59 $\pm$ 0.10} & 69.69 $\pm$ 7.48 & \textbf{94.13 $\pm$ 0.94} \\ 
                             & 50                      & \textbf{99.90 $\pm$ 0.02} & \textbf{99.92 $\pm$ 0.01} & \textbf{82.14 $\pm$ 1.18} & \textbf{89.69 $\pm$ 2.49} & 99.61 $\pm$ 0.19 & \textbf{99.72 $\pm$ 0.05} & \textbf{77.15 $\pm$ 2.89} & 92.91 $\pm$ 1.43 \\  \bottomrule 
                             
\multirow{6}{*}{hist}       & 100                     & 83.01 $\pm$ 0.60 & 80.92 $\pm$ 0.18 & 63.78 $\pm$ 2.37 & 73.64 $\pm$ 2.14 & 88.00 $\pm$ 0.22 & 77.31 $\pm$ 0.27 & 63.40 $\pm$ 1.06 & 83.56 $\pm$ 2.13 \\ \cline{2-10} 
                             & 90                      & 89.27 $\pm$ 0.41 & 84.50 $\pm$ 0.80 & 68.93 $\pm$ 0.88 & 77.51 $\pm$ 0.72 & 92.46 $\pm$ 0.57 & 79.94 $\pm$ 0.83 & 66.35 $\pm$ 1.82 & 86.34 $\pm$ 1.15 \\ 
                             & 80                      & 94.43 $\pm$ 0.67 & 87.24 $\pm$ 0.50 & 72.87 $\pm$ 2.63 & 77.12 $\pm$ 2.51 & 95.50 $\pm$ 0.67 & 83.49 $\pm$ 0.59 & 69.23 $\pm$ 1.06 & 85.54 $\pm$ 2.00 \\ 
                             & 70                      & 95.23 $\pm$ 6.09 & 89.86 $\pm$ 2.21 & 72.42 $\pm$ 4.33 & 76.69 $\pm$ 5.84 & \underline{97.35 $\pm$ 4.18} & 85.97 $\pm$ 2.10 & \underline{72.67 $\pm$ 1.16} & \textbf{92.43 $\pm$ 1.03} \\ 
                             & 60                      & \underline{96.25 $\pm$ 6.11} & \textbf{91.55 $\pm$ 2.31} & \underline{74.73 $\pm$ 1.93} & \underline{83.84 $\pm$ 4.56} & \textbf{97.87 $\pm$ 4.38} & \underline{87.94 $\pm$ 2.70} & 70.61 $\pm$ 1.15 & \underline{92.35 $\pm$ 1.68} \\ 
                             & 50                      & \textbf{96.81 $\pm$ 6.82} & \underline{90.91 $\pm$ 2.69} & \textbf{75.39 $\pm$ 1.98} & \textbf{86.68 $\pm$ 0.89} & 95.94 $\pm$ 5.50 & \textbf{89.17 $\pm$ 2.65} & \textbf{76.47 $\pm$ 3.75} & 92.23 $\pm$ 1.42 \\  \bottomrule 
                             
\multirow{6}{*}{ind}        & 100                     & 82.76 $\pm$ 0.71 & 84.58 $\pm$ 0.33 & 66.06 $\pm$ 2.26 & 70.93 $\pm$ 2.25 & 84.93 $\pm$ 0.27 & 82.31 $\pm$ 0.24 & 67.06 $\pm$ 0.81 & 80.88 $\pm$ 1.28 \\ \cline{2-10} 
                             & 90                      & 88.18 $\pm$ 0.59 & 89.34 $\pm$ 0.55 & 70.75 $\pm$ 0.93 & 75.64 $\pm$ 2.36 & 89.76 $\pm$ 0.39 & 84.96 $\pm$ 0.91 & 69.81 $\pm$ 1.52 & 83.37 $\pm$ 1.56 \\ 
                             & 80                      & 93.18 $\pm$ 0.53 & 92.79 $\pm$ 0.93 & 75.36 $\pm$ 2.52 & 73.70 $\pm$ 5.28 & 93.81 $\pm$ 0.37 & 88.71 $\pm$ 0.33 & 73.27 $\pm$ 1.02 & 82.50 $\pm$ 2.33 \\ 
                             & 70                      & 95.14 $\pm$ 5.94 & 94.59 $\pm$ 4.28 & 75.28 $\pm$ 3.59 & 74.17 $\pm$ 5.77 & \underline{95.52 $\pm$ 5.89} & 91.16 $\pm$ 2.90 & 77.07 $\pm$ 0.54 & 90.82 $\pm$ 0.85 \\ 
                             & 60                      & \underline{96.07 $\pm$ 6.15} & \textbf{96.36 $\pm$ 4.84} & \textbf{77.08 $\pm$ 1.77} & \underline{82.36 $\pm$ 4.60} & \textbf{96.78 $\pm$ 6.04} & \underline{93.10 $\pm$ 3.57} & \underline{77.74 $\pm$ 1.66} & \underline{91.40 $\pm$ 1.37} \\ 
                             & 50                      & \textbf{96.80 $\pm$ 6.84} & \underline{94.63 $\pm$ 5.72} & \underline{76.93 $\pm$ 1.37} & \textbf{85.01 $\pm$ 2.03} & 92.49 $\pm$ 9.71 & \textbf{94.31 $\pm$ 3.73} & \textbf{81.14 $\pm$ 4.06} & \textbf{92.10 $\pm$ 1.31} \\ \bottomrule  
\end{tabular}}
\end{table*}

\begin{table*}[!htb]
\caption{AP and AUC score for transductive dynamic link prediction with \textbf{DyGFormer} with random, historical, and inductive negative sampling strategies for various coverage rates. The best and second-best performing results are emphasized by \textbf{bold} and \underline{underlined} fonts.}
\label{tab:Trans_AP_AUC}
\centering
\resizebox{\linewidth}{!}{
\begin{tabular}{c|c|cccc|cccc}
\hline
\multirow{2}{*}{NSS}         & \multirow{2}{*}{Coverage} & \multicolumn{4}{c|}{AP}                                               & \multicolumn{4}{c}{AUC}                                        \\ \cline{3-10}
                             &       (\%)                   & Wikipedia       & Reddit          & UN Trade         & Can. Parl.      & Wikipedia       & Reddit          & UN Trade         & Can. Parl.      \\ \hline
                             \hline

\multirow{6}{*}{rnd}  & 100                       & 99.11 $\pm$ 00.02               & 99.21 $\pm$ 00.01            & 65.37 $\pm$ 00.29              & 98.32 $\pm$ 00.19                 & 99.01 $\pm$ 00.01               & 99.14 $\pm$ 00.01            & 69.06 $\pm$ 00.37              & 98.01 $\pm$ 00.18                \\ \cline{2-10} 
                      & 90                        & 99.60 $\pm$ 00.03               & 99.64 $\pm$ 00.05            & 61.02 $\pm$ 05.45              & 99.04 $\pm$ 00.48                 & 99.57 $\pm$ 00.04               & 99.62 $\pm$ 00.06            & 64.72 $\pm$ 03.46              & 98.81 $\pm$ 00.62                \\
                      & 80                        & 99.86 $\pm$ 00.04               & 99.78 $\pm$ 00.05            & \textbf{67.45 $\pm$ 07.22}& 99.32 $\pm$ 00.22                 & 99.86 $\pm$ 00.07               & 99.81 $\pm$ 00.09            & \underline{69.30 $\pm$ 03.73}& 99.10 $\pm$ 00.41                \\
                      & 70                        & 99.86 $\pm$ 00.05               & \underline{99.87 $\pm$ 00.02}      & \textbf{ 77.76 $\pm$ 01.59}& 99.05 $\pm$ 00.44                 & 99.92 $\pm$ 00.05               & \underline{99.93 $\pm$ 00.01}      & \textbf{75.45$\pm$ 0.03}& 98.49 $\pm$ 00.81                \\
                      & 60                        & \textbf{99.93 $\pm$ 00.01}      & \textbf{99.88 $\pm$ 00.01}   & {62.40 $\pm$ 11.67}     & \underline{99.52 $\pm$ 00.52}           & \textbf{99.96 $\pm$ 00.00}      & \textbf{99.94 $\pm$ 00.01}   & 65.46 $\pm$ 09.23              & \underline{99.31 $\pm$ 00.83}          \\
                      & 50                        & \underline{99.92 $\pm$ 00.02}         & 99.13 $\pm$ 01.11            & 60.21 $\pm$ 03.03              & \textbf{99.72 $\pm$ 00.23}        & \underline{99.96 $\pm$ 00.01}         & 99.65 $\pm$ 00.42            & {66.19 $\pm$ 04.06}     & \textbf{99.67 $\pm$ 00.27}       \\ \bottomrule
                      
\multirow{6}{*}{hist} & 100                       & 74.60 $\pm$ 05.62               & 81.39 $\pm$ 01.49            & 60.73 $\pm$ 02.84              & 98.57 $\pm$ 00.12                 & 75.00 $\pm$ 02.99               & 80.56 $\pm$ 00.46            & 70.70 $\pm$ 03.06              & 98.28 $\pm$ 00.18                \\ \cline{2-10} 
                      & 90                        & 74.39 $\pm$ 03.53               & 84.42 $\pm$ 02.18            & 58.81 $\pm$ 02.81              & 98.97 $\pm$ 00.70                 & 78.67 $\pm$ 02.87               & 86.48 $\pm$ 01.20            & \underline{63.94 $\pm$ 03.04}        & 98.68 $\pm$ 00.93                \\
                      & 80                        & \textbf{82.89 $\pm$ 08.86}      & 86.12 $\pm$ 02.04            & \underline{70.69 $\pm$ 08.71}& 99.33 $\pm$ 00.28                 & \underline{85.53 $\pm$ 06.40}         & 90.18 $\pm$ 01.19            & \textbf{71.99 $\pm$ 05.76}& 99.09 $\pm$ 00.48                \\
                      & 70                        & 80.86 $\pm$ 08.52               & 87.13 $\pm$ 00.92            & \textbf{80.53$\pm$ 7.57}& 98.97 $\pm$ 00.47                 & \textbf{85.89 $\pm$ 07.03}      & \underline{92.21 $\pm$ 00.37}      & 75.74 $\pm$ 9.98& 98.31 $\pm$ 00.91                \\
                      & 60                        & 71.71 $\pm$ 16.94               & \underline{88.35 $\pm$ 02.93}      & 57.65 $\pm$ 10.77              & \underline{99.47 $\pm$ 00.55}           & 79.09 $\pm$ 15.62               & \textbf{92.99 $\pm$ 01.11}   & 58.43 $\pm$ 09.18              & \underline{99.22 $\pm$ 00.90}          \\
                      & 50                        & \underline{77.36 $\pm$ 21.58}         & \textbf{89.47 $\pm$ 01.96}   & 58.31 $\pm$ 04.63              & \textbf{99.67 $\pm$ 00.28}        & 82.18 $\pm$ 23.39               & 89.45 $\pm$ 05.38            & 62.83 $\pm$ 06.28              & \textbf{99.60 $\pm$ 00.33}       \\ \bottomrule
                      
\multirow{6}{*}{ind}  & 100                       & 67.87 $\pm$ 12.41               & 91.17 $\pm$ 00.79            & 56.47 $\pm$ 01.79              & 98.66 $\pm$ 00.10                 & 69.88 $\pm$ 07.37               & 86.52 $\pm$ 00.82            & 65.25 $\pm$ 02.23              & 98.37 $\pm$ 00.16                \\ \cline{2-10} 
                      & 90                        & 64.93 $\pm$ 06.60               & 95.11 $\pm$ 00.92            & 60.74 $\pm$ 03.45              & 99.20 $\pm$ 00.46                 & 71.65 $\pm$ 05.05               & 93.89 $\pm$ 00.85            & 66.49 $\pm$ 03.17              & 99.00 $\pm$ 00.60                \\
                      & 80                        & \underline{76.77 $\pm$ 09.48}         & 97.72 $\pm$ 00.95            & \underline{72.23 $\pm$ 08.04}& 99.45 $\pm$ 00.29                 & \textbf{80.53 $\pm$ 07.03}      & 97.92 $\pm$ 01.19            & \underline{74.00 $\pm$ 05.45}    & 99.26 $\pm$ 00.48                \\
                      & 70                        & 72.54 $\pm$ 15.14               & \underline{98.72 $\pm$ 00.81}      & \textbf{82.98$\pm$ 6.71}& 99.05 $\pm$ 00.45                 & \underline{78.35 $\pm$ 13.70}         & \underline{99.25 $\pm$ 00.39}      & 79.65$\pm$ 06.71& 98.50 $\pm$ 00.83                \\
                      & 60                        & 68.53 $\pm$ 21.58               & \textbf{99.38 $\pm$ 00.38}   & 64.11 $\pm$ 11.79              & \underline{99.53 $\pm$ 00.53}           & 73.34 $\pm$ 20.90               & \textbf{99.64 $\pm$ 00.14}   & 67.77 $\pm$ 09.98              & \underline{99.33 $\pm$ 00.85}          \\
                      & 50                        & \textbf{77.02 $\pm$ 28.63}      & 95.00 $\pm$ 05.90            & {69.00 $\pm$ 05.72}     & \textbf{99.74 $\pm$ 00.21}        & 77.23 $\pm$ 31.64               & 92.84 $\pm$ 09.24            & \textbf{75.88 $\pm$ 05.65}     & \textbf{99.69 $\pm$ 00.26}       \\ \bottomrule
\end{tabular}}
\end{table*}

\begin{table*}[!htb]
\caption{AP and AUC score for inductive dynamic link prediction by \textbf{DyGFormer} with random, historical, and inductive negative sampling strategies for various coverage rates. The best and second-best performing results are emphasized by \textbf{bold} and \underline{underlined} fonts.}
\vspace{1mm}
\label{tab:Ind_AP_AUC}
\centering
\resizebox{\linewidth}{!}{
\begin{tabular}{c|c|cccc|cccc}
\hline
\multirow{2}{*}{NSS}  & \multirow{2}{*}{Coverage} & \multicolumn{4}{c|}{AP}                                                                                           & \multicolumn{4}{c}{AUC}                                                                                          \\ \cline{3-10} 
                      &           (\%)            & {Wikipedia} & {Reddit} & {UN Trade} & {Can. Parl.} & {Wikipedia} & {Reddit} & {UN Trade} & {Can. Parl.} \\ \hline
                             \hline
\multirow{6}{*}{rnd}  & 100                       & 98.66 ± 00.04               & 98.73 ± 00.03            & 63.03 ± 00.26              & 93.13 ± 00.57                 & 98.57 ± 00.02               & 98.58 ± 00.03            & 65.90 ± 00.32              & 91.36 ± 00.86                \\ \cline{2-10} 
                      & 90                        & 99.38 ± 00.06               & 99.32 ± 00.08            & 59.56 ± 04.65              & 94.73 ± 00.87                 & 99.35 ± 00.04               & 99.24 ± 00.09            & 62.68 ± 02.80              & 92.59 ± 00.69                \\
                      & 80                        & 99.77 ± 00.04               & 99.66 ± 00.11            & \underline{66.13 ± 08.17}  & 95.93 ± 00.88                 & 99.81 ± 00.07               & 99.66 ± 00.16            & \underline{67.06± 04.76}   & 94.09 ± 01.24                \\
                      & 70                        & \underline{99.85 ± 00.05}         & \underline{99.81 ± 00.01}      &\textbf{ 76.05 ± 1.44}& 96.15 ± 00.22                 & 99.89 ± 00.07               & \underline{99.89 ± 00.01}      & { \textbf{73.26 ± 2.64}}& 94.26 ± 00.29                \\
                      & 60                        & 99.82 ± 00.04               & \textbf{99.81 ± 00.01}   & 60.29 ± 11.96              & \underline{96.76 ± 01.56}           & \underline{99.90 ± 00.02}         & \textbf{99.89 ± 00.01}   & 62.11 ± 09.78              & \underline{95.46 ± 02.24}          \\
                      & 50                        & \textbf{99.85 ± 00.02}      & 98.76 ± 01.44            & 56.92 ± 02.35              & \textbf{97.24 ± 01.56}        & \textbf{99.92 ± 00.01}      & 99.44 ± 00.66            & 61.56 ± 03.49              & \textbf{95.97 ± 02.34}       \\ \bottomrule
                      
\multirow{6}{*}{hist} & 100                       & 63.13 ± 09.13               & 61.80 ± 01.32            & 53.48 ± 01.52              & 93.41 ± 00.54                 & 64.28 ± 06.05               & 62.78 ± 00.26            & 60.61 ± 02.21              & 91.79 ± 00.72                \\ \cline{2-10} 
                      & 90                        & 61.22 ± 05.53               & 63.41 ± 01.09            & 58.02 ± 04.60              & 95.05 ± 00.88                 & 65.74 ± 04.76               & 65.75 ± 00.30            & 62.86 ± 03.68              & 93.09 ± 00.65                \\
                      & 80                        & \textbf{69.87 ± 06.59}      & 63.68 ± 00.98            & \underline{65.44 ± 06.97}& 96.29 ± 00.68                 & \textbf{73.77 ± 05.50}      & 67.94 ± 00.53            & { 66.95 ± 04.56}        & 94.54 ± 01.04                \\
                      & 70                        & \underline{67.35 ± 12.76}         & 64.18 ± 01.47            & \textbf{79.83 ± 6.85}& 96.31 ± 00.27                 & \underline{71.72 ± 12.03}         & \textbf{70.39 ± 00.87}   & \textbf{75.73 ± 9.28}& 94.43 ± 00.39                \\
                      & 60                        & 61.81 ± 14.01               & \underline{65.11 ± 01.81}      & 59.21 ± 10.25              & \underline{96.89 ± 01.42}           & 67.78 ± 16.27               & \underline{70.18 ± 02.23}      & 61.19 ± 08.50              & \underline{95.60 ± 02.11}          \\
                      & 50                        & 66.38 ± 19.47               & \textbf{69.21 ± 04.52}   & 61.33 ± 03.75              & \textbf{97.35 ± 01.50}        & 70.91 ± 26.14               & 69.35 ± 06.52            & \underline{67.26 ± 04.67}     & \textbf{96.14 ± 02.24}       \\ \bottomrule
                      
\multirow{6}{*}{ind}  & 100                       & 63.13 ± 09.15               & 61.77 ± 01.32            & 53.51 ± 01.51              & 93.42 ± 00.54                 & 64.28 ± 06.05               & 62.76 ± 00.26            & 60.65 ± 02.20              & 91.81 ± 00.72                \\ \cline{2-10} 
                      & 90                        & 61.19 ± 05.52               & 63.38 ± 01.09            & 58.04 ± 04.59              & 95.06 ± 00.88                 & 65.72 ± 04.77               & 65.73 ± 00.30            & 62.88 ± 03.66              & 93.10 ± 00.65                \\
                      & 80                        & \textbf{69.86 ± 06.60}      & 63.66 ± 00.97            & \underline{68.95 ± 06.27}  & 96.30 ± 00.68                 & \textbf{73.76 ± 05.49}      & 67.92 ± 00.53            & \underline{69.82 ± 04.54}  & 94.55 ± 01.04                \\
                      & 70                        & \underline{67.34 ± 12.78}         & 64.14 ± 01.46            & \textbf{79.85 ± 6.86}& 96.32 ± 00.27                 & \underline{71.72 ± 12.04}         & \textbf{70.36 ± 00.86}   & \textbf{75.76 ± 9.28}& 94.43 ± 00.39                \\
                      & 60                        & 61.80 ± 14.02               & \underline{65.09 ± 01.80}      & 59.23 ± 10.25              & \underline{96.90 ± 01.41}           & 67.78 ± 16.27               & \underline{70.15 ± 02.24}      & 61.23 ± 08.51              & \underline{95.61 ± 02.11}          \\
                      & 50                        & 66.40 ± 19.52               & \textbf{69.19 ± 04.50}   & 61.34 ± 03.75              & \textbf{97.35 ± 01.51}        & 70.92 ± 26.20               & 69.33 ± 06.48            & {67.27 ± 04.66}     & \textbf{96.13 ± 02.25}       \\ \bottomrule
\end{tabular}}
\end{table*}

\begin{table*}[!htb]
\centering
\caption{AUC for coverage-based dynamic node classification by \textbf{DyGFormer} for various coverage rates, with and without handling class imbalance. The best and second-best performing results are emphasized in \textbf{bold} and \underline{underlined} fonts, respectively.}
\label{tab:dygformer_node_clf}
\resizebox{0.4\linewidth}{!}{
\begin{tabular}{c|c|cc}
\hline
\multirow{2}{*}{Coverage} & \multirow{2}{*}{$\beta$} & \multicolumn{2}{c}{DyGFormer}                           \\ \cline{3-4} 
                    (\%)      &                          & {Wikipedia  }           \\ \bottomrule
100                       & $=1$                     & {87.04 $\pm$ 1.08} \\ \hline
\multirow{2}{*}{90}       & $=1$                     & {87.19 $\pm$ 0.86} \\ 
                          & $=2$                     & {88.02 $\pm$ 1.87} \\ \hline
\multirow{2}{*}{80}       & $=1$                     & {88.47 $\pm$ 3.32} \\ 
                          & $=2$                     & {\underline{88.97 $\pm$ 2.37}} \\ \hline
\multirow{2}{*}{70}       & $=1$                     & {86.96 $\pm$ 3.60} \\ 
                          & $=2$                     & {\textbf{89.17 $\pm$ 4.72}} \\ \hline
\multirow{2}{*}{60}       & $=1$                     & {83.79 $\pm$ 3.76} \\ 
                          & $=2$                     & {83.42 $\pm$ 2.71} \\ \bottomrule
\end{tabular}}
\end{table*}

\end{document}